\documentclass[a4paper,fleqn]{cas-sc}

\usepackage[numbers]{natbib}
\usepackage{caption,subcaption}
\usepackage{bbm} 
\usepackage{comment}

\makeatletter
\newcommand\incircbin
{%
  \mathpalette\@incircbin
}
\newcommand\@incircbin[2]
{%
  \mathbin%
  {%
    \ooalign{\hidewidth$#1#2$\hidewidth\crcr$#1\bigcirc$}%
  }%
}

\makeatother

\def\tsc#1{\csdef{#1}{\textsc{\lowercase{#1}}\xspace}}
\tsc{EK}
\tsc{SR}

\begin{document}
\let\WriteBookmarks\relax
\def\floatpagepagefraction{1}
\def\textpagefraction{.001}
\shorttitle{Lower Dimensional Kernels for Video Discriminators}
\shortauthors{E Kahembwe et~al.}

\title [mode = title]{Lower Dimensional Kernels for Video Discriminators}
\title [mode=alt]{Lower-Dimensional Video Discriminators for Generative Adversarial Networks}

\author[1,2,3]{Emmanuel Kahembwe}[type=editor,
                        auid=000,bioid=1, 
                        orcid=0000-0003-2064-2134]
\cormark[1]
\ead{e.kahembwe@ed.ac.uk}

\author[1,2,4]{Subramanian Ramamoorthy}
\address[1]{Robust Autonomy and Decisions Group, The School of Informatics, The University of Edinburgh, 10 Crichton St, Edinburgh EH8 9AB}
\address[2]{The Edinburgh Centre of Robotics, The University of Edinburgh's Bayes Centre, 47 Potterrow,
Edinburgh EH8 9BT}
\address[3]{The School of Engineering and Physical Sciences, The Robotarium, Heriot-Watt University,  Edinburgh, EH14 4AS}
\address[4]{FiveAI, 5th Floor, Greenside, 12 Blenheim Place, Edinburgh, EH7 5JH}

\cortext[cor1]{Corresponding author.}

\nonumnote{This research is supported by the Engineering and Physical Sciences Research Council (EPSRC), as part
of the CDT in Robotics and Autonomous Systems at Heriot-Watt University and The University of Edinburgh.
  }

\begin{abstract}
This work presents an analysis of the discriminators used in Generative Adversarial Networks (GANs) for Video. We show that unconstrained video discriminator architectures induce a loss surface with high curvature which make optimisation difficult. We also show that this curvature becomes more extreme as the maximal kernel dimension of video discriminators increases. With these observations in hand, we propose a family of efficient Lower-Dimensional Video Discriminators for GANs (LDVD GANs). The proposed family of discriminators improve the performance of video GAN models they are applied to and demonstrate good performance on complex and diverse datasets such as UCF-101. In particular, we show that they can double the performance of Temporal-GANs and provide for state-of-the-art performance on a single GPU.
\end{abstract}


\maketitle

\section{Introduction}

\begin{figure}[pos=!h]
    \centering
    \includegraphics[width=1\linewidth]{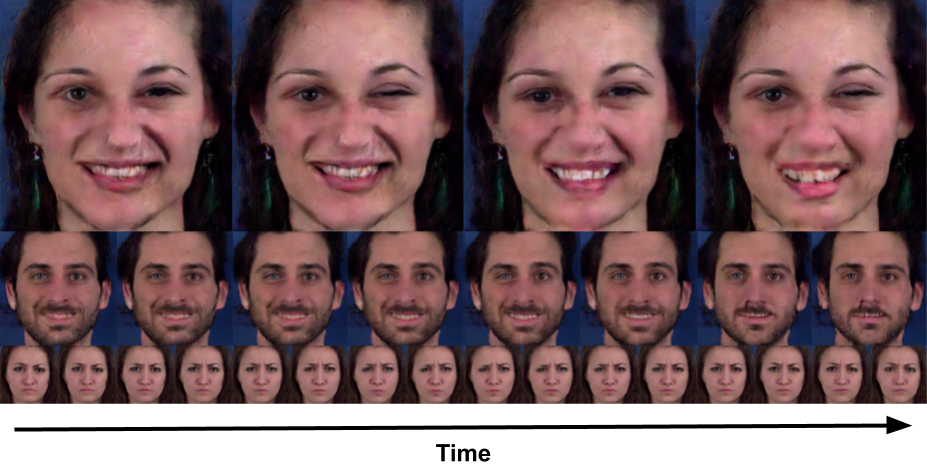}
    
    \caption{Frames from videos generated by our TGAN-F model trained on MUG-FED at 256$\times$256,
128$\times$128, and 64$\times$64 resolutions (top to bottom).} 
    \label{fig:MUG_ldvd_64_256}
\end{figure}

\subsection{Motivation}
\citet{GoodfellowPMXWOCB14generative} introduced generative adversarial training of neural networks as a way to model complex data distributions. They demonstrated the efficacy of this training regime on the task of image generation, subsequently establishing the field of Generative Adversarial Networks (GANs). GAN-based image generation has since observed significant advances; from architectural contributions \cite{RadfordMC15unsupervised,zhangGMO2018attention, karras2018progressive} to novel forms of losses \cite{ArjovskyCB17wasserstein,jolicoeur-martineau2018relativistic} and stabilization methods \cite{MiyatoKKY18spectral,HeuselRUNH17updaterule, salimansGZCRCC2016improved,Gulrajani2017improved}. Current state-of-the-art models for image generation produce high-resolution visual results that are sometimes difficult for humans to distinguish from those derived from the true data distribution \cite{brockDS2018large,karrasLA2018style}.

In comparison, video generation has not enjoyed the same level of progress as image generation. Video GAN (VGAN) models have benefited from methods such as progressive growing \cite{AcharyaHPG2018towards} and Lipschitz regulation \cite{SaitoMS17temporal}. However, VGAN models are still insufficient when it comes to modelling the true video distribution and produce results that are easily identified as lying outside its support.

There are many possible reasons for the comparatively limited performance of GAN models on the task of video modelling when compared to image modelling. The most obvious reason being that video modelling is a higher dimensional and more complex task due to the addition of a temporal dimension. The additional dimension significantly increases the number of parameters required by a GAN model to sufficiently capture the true data distribution, resulting in higher memory and compute costs. 

\subsection{Summary}
We analyse the discriminators used in VGAN models and reveal a more nuanced contributing factor to the limited performance of these models. We find that the dimensionality of the 3D kernels used in video discriminators induces higher curvature in the loss landscape and that this is detrimental to first order optimization methods such as stochastic gradient decent. In light of this observation, we target the maximum kernel dimensionality of video GAN models as an area of optimisation for the purposes of stabilizing training and improving performance. We also explore computation and memory efficiency from this perspective. To conclude, we demonstrate that in lowering the maximal kernel dimensionality of a video GAN model, we can reduce pathologies in the loss landscape, improve overall model performance while significantly increasing memory and computation efficiency.

\subsection{Layout}
This work is organized as follows; in the following section we review the relevant literature in both the image and video GAN domains. Section~\ref{sec:disc_arch} details an analysis of the MoCoGAN and TGAN video discriminators. Section~\ref{sec:ldvd} introduces a family of lower dimensional video discriminators. Secion~\ref{sec:experiments} compares the performance of the different discriminators against prior art. Section~\ref{sec:end} concludes this study with a discussion of the results and its implications for the design of video discriminators.

\section{Related Work}
\label{sec:related}
\subsection{Image Generation}
Adversarial training, within the context of neural networks \cite{GoodfellowPMXWOCB14generative}, pits two networks against each other in a zero-sum non-cooperative game which is solved, in the game-theoretic sense, by the application of the minimax theorem \cite{vNeumann1928}. A network, called the generator ($G$), learns to model the true data distribution by fooling an adversary, termed the discriminator ($D$), whose job is to learn a classifier that tells apart the generated data from the real data. The standard formulation of this game is given by the value function $V(G, D)$;
\begin{equation}
    \underset{\theta}{\text{min }}  
    \underset{\psi}{\text{max }} V(G, D) = \mathbbm{E}_{x \sim  p_{data}(x)}\bigg[f_D\bigg(D_\psi(x)\bigg)\bigg] + 
    \mathbbm{E}_{z \sim  p_z(z)}\bigg[f_G\bigg(D_\psi\Big(G_\theta(z)\Big)\bigg)\bigg]
    \label{eq:gan_minmax}
\end{equation}

where $f(\cdot)$ is a real-valued function whose exact form depends on the choice of loss function \cite{GoodfellowPMXWOCB14generative, ArjovskyCB17wasserstein,Nowozin16FGAN}.
\citet{GoodfellowPMXWOCB14generative} initially presented models trained with this approach, termed Generative Adversarial Networks (GANs), on the task of image generation. In the GAN research community, significant effort has been dedicated to improving performance on image generation tasks as a benchmark and to stabilizing the adversarial training regime \cite{,RadfordMC15unsupervised,ArjovskyCB17wasserstein, MiyatoKKY18spectral, zhangGMO2018attention, brockDS2018large}. To facilitate evaluation of GAN research, metrics such as the Inception Score (IS) \cite{salimansGZCRCC2016improved} and the Fr\'echet Inception Distance (FID) \cite{HeuselRUNH17updaterule} have become the standard for benchmarking image quality and diversity respectively. 

\subsection{Video Generation}
Generative adversarial networks for video (VGAN) by \citet{VondrickPT16generating} was the first model to extend GANs to the video domain. It is composed of a 3D discriminator and a two-stream generator; one stream generating the static background content using 2D kernels and another 3D stream generating the foreground dynamic content. 

Subsequent models such as Temporal GAN (TGAN) \cite{SaitoMS17temporal} and Motion and Content decomposed GAN (MoCoGAN) \cite{Tulyakov0YK18mocogan} use a two-stage generator. In TGAN, the first stage samples a random latent $z_c \in \mathbb{R}^{d_c}$ and generates a set of latent variables $z_m^{1..T}$ conditioned on $z_c$ that define a video trajectory across $T$ time-steps. The second stage takes the concatenated latents $[z_c;z_m^t]$,\footnote{ $[z_c;z_m]$ denotes the concatenation operation   ;   between $z_c$ and $z_m$} and generates a single image for each time-step $t \in T$. The first stage of TGAN's generator uses 1D kernels, followed by 2D kernels for image generation in the second stage. It utilises 3D kernels in its discriminator.

MoCoGAN \cite{Tulyakov0YK18mocogan} explicitly models the static and dynamic attributes of video separately. This model assumes a factorized latent space, where the content in video is embedded on a subspace, $z_c \in \mathbb{R}^{d_c}$ and its associated motion is embedded on another subspace, $z_m \in \mathbb{R}^{d_m}$. The generator models the joint space $Z_l \in \mathbb{R}^d$, where $\mathbb{R}^d = \mathbb{R}^{d_m} + \mathbb{R}^{d_c}$, using a combination of a Gated Recurrent Unit (GRU) \cite{cho-etal-2014-learning} to generate the temporal dynamics of video and a 2D convolutional upscaling network to generate the associated spatial information. The video generation process involves sampling a latent variable $z_m^t$ at each time-step $t \in T$ from the motion subspace, processing it with a GRU and concatenating the resulting output with $z_c$ to form $z = [z_c;GRU(z_m^{1-T})]$, where $z \in Z_l$. The upscaling network is then used to project $z$ to the image space to form video frames. The MoCoGAN discriminator architecture consists of two components, a 2D image and 3D video discriminator. It is the current state-of-the-art model for low-resolution, 64x64 video generation \cite{soomroZS2012ucf101}.\footnote{We refer to the state-of-the-art as benchmarked on the UCF-101 dataset}

In the high-resolution video GAN literature, \citet{AcharyaHPG2018towards} propose Progressive Video GAN (ProVGAN), a model that combines progressive growing with 3D kernels and a sliced-wasserstein loss to generate video at 256x256 resolution. \citet{saito2018tganv2} explore sub-sampling as a way to scale video GAN models to 192x192 video generation and achieve state-of-the-art unconditional generation results when combined with a large batch training regime. Their proposed model, TGANv2, reduces memory and computational costs by sub-sampling across time, space and batch size as the resolution of feature maps increases. The TGANv2 video generator is comprised of a convolutional LSTM and a 2D image generator. It also uses multiple rendering layers at different resolutions within the generation pipeline, in conjunction with a hierarchy of 3D discriminators to critic the generated video at different spatial and temporal resolutions.

\section{Discriminator Architectures}
\label{sec:disc_arch}
Although there is some variety in the architectures of video GAN generators, the associated discriminator architectures have remained fairly consistent. There are currently two primary choices for video GAN discriminators; a dual (2D image + 3D video) discriminator architecture as in MoCoGAN, or a 3D video discriminator architecture as in VGAN, TGAN and ProVGAN.\footnote{The dimensionality of the discriminator is always in reference to the maximum convolution kernel dimension}

In this section, we take a closer look at the current choice of video GAN discriminators and study the impact that architectural decisions for this component have on model performance. In particular, we explore the question;

\begin{itemize}
    \item \textit{"What makes a good discriminator for video GANs?"}.
\end{itemize}

To answer this question, we analyse the properties of seminal video GAN discriminators for both dual and single discriminator architectures. For the dual discriminator architecture, our analysis focuses on the MoCoGAN model since it is the first video generation model to incorporate multiple architectural components in its discriminator. TGAN is used as the representative model for single component discriminators due to the  architectural similarity of its video discriminator to that of MoCoGAN.

\newpage
\subsection{Experimental Setup}
The original experimental code for the MoCoGAN \footnote{MoCoGAN Code: https://github.com/sergeytulyakov/mocogan} and TGAN \footnote{TGAN Code: https://github.com/pfnet-research/tgan} model is publicly available.
As such, all experiments use the original experimental code and settings to allow for accurate analysis and aid reproducibility. 
Training and performance benchmarking is carried out on a single 12GB Titan-X GPU.

\subsubsection{Quantitative Experiments}
We use the UCF-101 dataset for our quantitative experiments \cite{soomroZS2012ucf101}. This dataset was initially introduced for action recognition but was co-opted by the GAN research field in order to benchmark video GAN models. The UCF-101 dataset \cite{soomroZS2012ucf101} is a video dataset consisting of 13,320 video clips divided across 101 different action categories, at a spatial resolution of 320 by 240 pixels.

Our preprocessing pipeline center crops all videos to 240x240 pixels. For the MoCoGAN experiments, the video is temporally subsampled by a factor of 2 and 16 consecutive frames are randomly extracted. For the TGAN experiments, the video is not subsampled in order to match the original experimental conditions.\footnote{Results for TGAN trained with temporal subsampling and MoCoGAN trained without temporal subsampling are explicitly labelled} We then resize the resulting video to the model resolution (i.e. 16x64x64 or 16x128x128). We use the \textit{"trainlist01"} training split containing 9537 videos to train our models as in previous works.

\paragraph{Evaluation Metrics} We benchmark performance against the video extensions of the Inception Score (IS) and Fr\'echet Inception Distance (FID) where a higher IS implies that the generated video is of a higher visual quality and a lower FID implies a better fit to the modes of the data distribution (i.e. good diversity).\footnote{These metrics are defined relatively and should be approached with caution \cite{barratt2018note}} These metrics are highly sensitive to implementation so we use their exact implementation from the original TGAN experiments \cite{SaitoMS17temporal}. We calculate the IS using the Sport1M pre-trained C3D classification model fine-tuned on UCF-101 from the original TGAN experiments. We calculate the FID using the activations from the second to last linear layer, denoted \textit{fc7}, of the same C3D model.
 
As in previous works, we generate 10000 samples from the model to evaluate each metric and derive a rough standard deviation by repeating this procedure four times. 

\subsubsection{Qualitative Results}
We use the MUG Facial Expression Database (MUG-FED) for our qualitative experiments \cite{AifantiPD10mug}. This dataset is composed of 1462 video sequences of 86 people demonstrating 7 categories of facial expression; anger, fear, disgust, happiness, sadness, surprise and neutral. The videos are recorded at a resolution of 896x896 pixels and range between 40-180 frames.

\paragraph{Evaluation Metrics} For each model, 10000 video samples are generated and randomly sub-divided into 100 batches. All videos in each batch are tiled and aggregated into a single larger video. The tiled videos are presented, two at a time, to human participants for a side-by-side visual evaluation of sample quality and batch diversity.

\subsection{MoCoGAN Discriminator}
\label{sec:mocogan}
Although the seminal video GAN models used a single 3D video discriminator, later works such as MoCoGAN achieve better results with a dual (2D image + 3D video) discriminator model. In theory, a high-capacity discriminator with kernels whose dimensionality matches that of the input data distribution should allow for better criticism. In practice, this is not observed and the MoCoGAN authors attribute the performance boost gained with the addition of an image level discriminator to its ability to "focus on static appearances". In the following section, we investigate this claim and provide a more thorough explanation grounded with empirical observations.

\subsubsection{Ablation Study}
\label{sec:ablation}

Table \ref{table:mocoganDis} details the architectures for the MoCoGAN 2D image and 3D video discriminators. These are identical patch-level discriminators that operate on $46 \times 46$ patches of the input video frames with the distinction that the image discriminator is comprised of 2D convolution kernels and the video discriminator utilizes 3D kernels.

\begin{table}[width=.9\linewidth,cols=3,pos=!h]
\caption{MoCoGAN Image and Video Discriminators}
\label{table:mocoganDis}
\begin{tabular*}{\tblwidth}{@{} LLL@{} }
\toprule
Layer & Image Configuration &  Video Configuration \\
\midrule
Input & height $\times$ width $\times$ 3  & 16 $\times$ height $\times$ width $\times$ 3  \\
\midrule
c0 & Conv2D-(N64, K4, S2, P1), LReLU & Conv3D-(N64, K4, S(1,2,2), P(0,1,1)), LReLU\\
c1 & Conv2D-(N128, K4, S2, P1), BN, LReLU & Conv3D-(N128, K4, S(1,2,2), P(0,1,1)), BN, LReLU\\
c2 & Conv2D-(N256, K4, S2, P1), BN, LReLU & Conv3D-(N256, K4, S(1,2,2), P(0,1,1)), BN, LReLU\\
c3 & Conv2D-(N1, K4, S2, P1) & Conv3D-(N1, K4, S(1,2,2), P(0,1,1))\\
\bottomrule
\end{tabular*}
\end{table}

We replicate the results for this model on the UCF-101 dataset and ablate its discriminator components to ascertain how much each component contributes to the final model performance.  Results are presented in Table \ref{table:mocogan_ablate}. 

\begin{table}[width=.9\linewidth,cols=5,pos=!h]
\caption{MoCoGAN Ablation}
\label{table:mocogan_ablate}
\begin{tabular*}{\tblwidth}{@{} LLRCC@{} }
\toprule
Row & Model & Disc Params & IS  $\uparrow$ & FID  $\downarrow$ \\
\midrule
1 & MoCoGAN                     & 3.3M  & 11.58 $\pm$ .04   & 9485.34 $\pm$ 14.61   \\
2 & MoCoGAN - NoTemporalSubsampling  & 3.3M  & 10.35 $\pm$ .06 & 9657.44 $\pm$ 3.90  \\
\midrule
3 & MoCoGAN - Video Discriminator Only  & 2.7M  & 11.09 $\pm$ .03   & 9565.65 $\pm$ 21.03   \\
4 & MoCoGAN - Image Discriminator Only  & .7M  & 8.26 $\pm$ .04    & 10494.09 $\pm$ 12.52  \\
\midrule
5 & MoCoGAN - 2xChannels                & 13.2M & 12.61 $\pm$ .08   & 9166.81 $\pm$ 9.92    \\
6 & MoCoGAN - 2xChannels - Video Discriminator Only  & 10.5M  & 11.73 $\pm$ .05 & 9461.10 $\pm$ 1.85   \\
\midrule
7 & MoCoGAN - Video Discriminator Only - ksize2  & .3M   & 3.98 $\pm$ .02  & 13664.81 $\pm$ 7.64   \\
8 & MoCoGAN - Image Discriminator Only - ksize8  & 2.7M  & 7.62 $\pm$ .04  & 10337.71 $\pm$  0.47  \\
\midrule
9 & MoCoGAN- Original published in \cite{Tulyakov0YK18mocogan}      & 3.3M & 12.42 $\pm$ .03 &  \\
\bottomrule
\end{tabular*}
\end{table}

In Table~\ref{table:mocogan_ablate}, we observe that there is a degradation in performance without an image-level discriminator (row 1 vs row 3) and we also observe that image-level statistics can account for most of the model performance as demonstrated by the the model trained with an image-only discriminator (row 4). 

\begin{figure*}
    \centering
    \includegraphics[width=0.9\linewidth,height=10em]{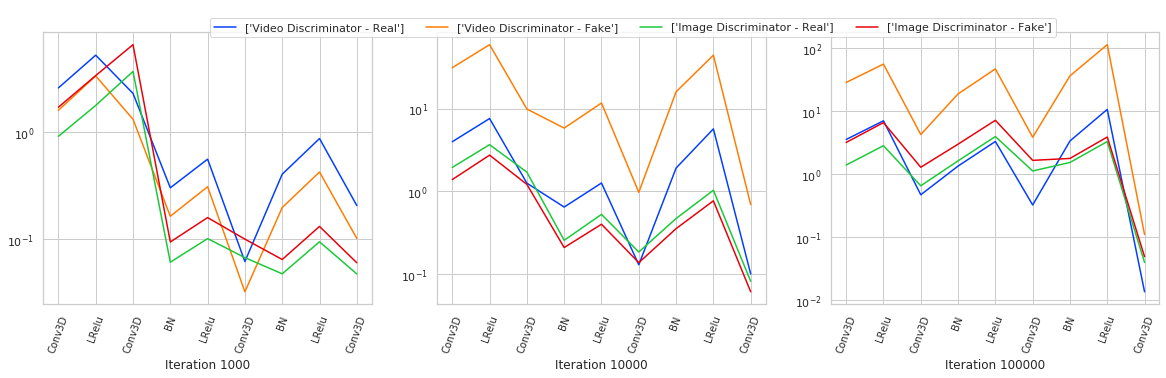}
    \caption{Norm of the gradient at each node in the computation graph for batches of real and fake data.} 
    \label{fig:MoCoGAN_GradNorm}
\end{figure*}

We were not able to replicate the published results and suspect that this could be due to the MoCoGANs saturating loss function. The loss function used in the original GAN formulation set $f_D(t) = \text{log}(t)$ and $f_G(t) = \text{log}(1 - t)$ in Equation \ref{eq:gan_minmax}. $f_G$ saturates when the discriminator overpowers the generator and learns a perfect classifier between the real and fake data distribution. As a result, the gradient signal propagated back to the generator via the discriminator vanishes, stalling training for the generator. \citet{GoodfellowPMXWOCB14generative} proposed a solution for this by training the generator to maximise $f_D$, i.e. $f_G = -f_D$. This provides for a non-saturating version of the GAN loss that allows for a non-vanishing gradient signal through out training (Figure \ref{fig:MoCoGAN_GradNorm}). The non-saturating loss did not improve performance but we maintain it for all further experiments to avoid potential saturation issues. We were only able to to replicate the published performance of the MoCoGAN model, by doubling the number of channels in every component of the model (row 5). Doubling just the channels for the video discriminator was not sufficient even though it accounts for 80\% of the discriminator parameters (row 6). Finally, we also measure the impact of temporal subsampling on model performance as it was not used in previous models such as TGAN (row 2). We observe that temporal subsampling accounts for a significant portion of model performance.

\subsubsection{Hessian Analysis}
\label{sec:hess}

\begin{figure}[pos=!th]
    \centering
    \begin{subfigure}[t]{0.5\textwidth}
        \centering
        \includegraphics[width=1\linewidth]{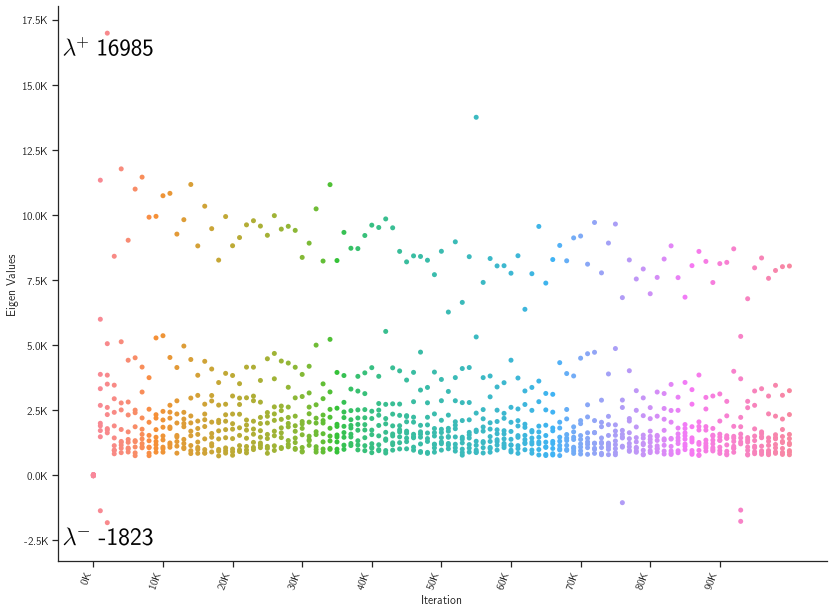}
        \caption{MoCoGAN- Image Discriminator}
        \label{fig:MoCoGAN_iDis_EV}
    \end{subfigure}%
    \begin{subfigure}[t]{0.5\textwidth}
        \centering
        \includegraphics[width=1\linewidth]{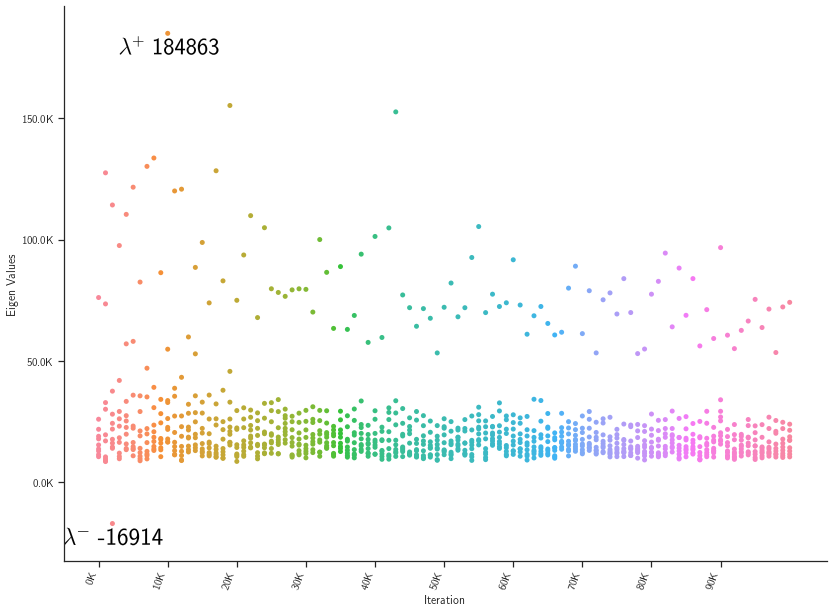}
        \caption{MoCoGAN- Video
        Discriminator}
        \label{fig:MoCoGAN_vDis_EV}
    \end{subfigure}%
    \\
    \begin{subfigure}[t]{0.5\textwidth}
        \centering
        \includegraphics[width=1\linewidth]{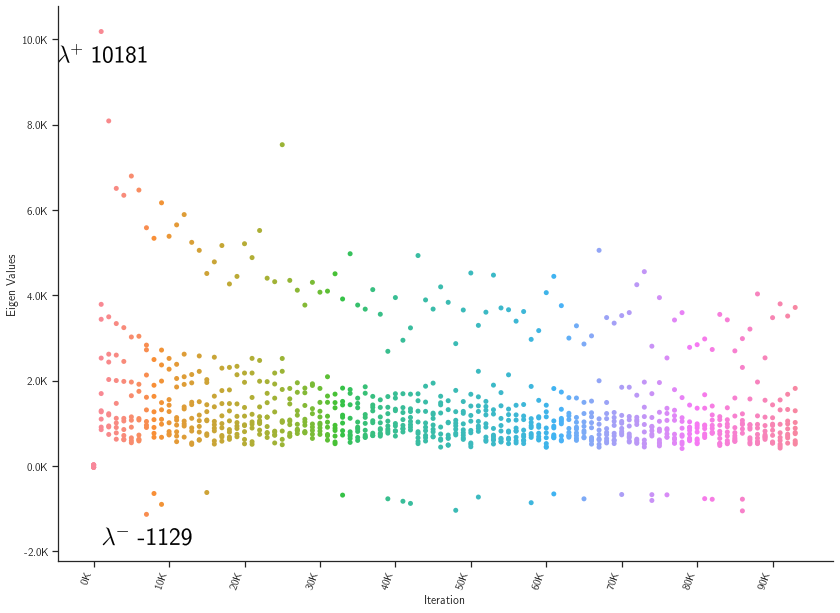}
        \caption{MoCoGAN 2xChannels - Image Discriminator}
        \label{fig:MoCoGAN_iDis_2xCh_EV}
    \end{subfigure}%
    \begin{subfigure}[t]{0.5\textwidth}
        \centering
        \includegraphics[width=1\linewidth]{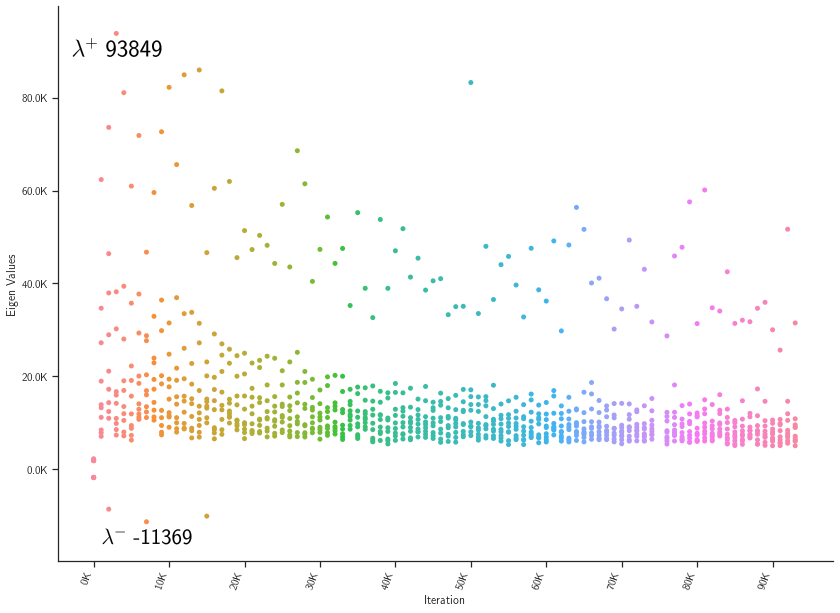}
        \caption{MoCoGAN 2xChannels - Video Discriminator}
        \label{fig:MoCoGAN_vDis_2xCh_EV}
    \end{subfigure}
\caption{The 10 largest eigenvalues of the loss Hessian with respect to the image and video discriminators of the MoCoGAN architecture. (a) and (b) show these values for the original MoCoGAN architecture. (c) and (d) show these for MoCoGAN with the number of channels doubled (see MocoGAN - 2xChannels in Table~\ref{table:mocogan_ablate}). $\lambda^{+}$ denotes the largest positive eigenvalue encountered during training and $\lambda^{-}$ the largest negative eigenvalue.} 
    \label{fig:MoCoGAN_Dis_EV}
\end{figure}

In this section, we analyse the loss surface induced by the image and video discriminators via analysis of the Hessian of the
GAN objective with respect to the discriminator parameters. Hessian analysis of neural networks is computationally expensive, especially for large models such as those used in the video GAN domain. As an approximation, we employ the $R$-operator from \citet{Pearlmutter94hess} to calculate the exact Hessian vector product for neural networks and combine it with the Lanczos algorithm to calculate the eigen spectra of the Hessian. The gradient and Hessian are taken with respect to the parameters of the discriminator and are given by
$\dot{\psi}  = \nabla V(G, D)$ and $ \ddot{\psi}  = \nabla^2 V(G, D) 
$ respectively. We track the leading eigenvalues of the Hessian throughout training and present these results in Figure \ref{fig:MoCoGAN_Dis_EV}.

Figure \ref{fig:MoCoGAN_Dis_EV} provides an interesting perspective as to what is going on with the MoCoGAN discriminator. The primary observation is that for identical discriminator architectures, an increase in kernel dimensionality results in a loss surface with significantly more pathological curvature. Each discriminator component individually encounters eigenvalues during training that are an order of magnitude larger than the majority of leading eigenvalues. But there is also a further order of magnitude difference between the maximum eigenvalue and majority leading eigenvalues for the 2D image discriminator, when compared against the 3D video discriminator (see Figure~\ref{fig:MoCoGAN_iDis_EV} vs \ref{fig:MoCoGAN_vDis_EV} and Figure~\ref{fig:MoCoGAN_iDis_2xCh_EV} vs \ref{fig:MoCoGAN_vDis_2xCh_EV} ). An increase in parameters results in a generally smoother loss landscape with possibly more saddle points (see Figure~\ref{fig:MoCoGAN_iDis_EV} vs \ref{fig:MoCoGAN_iDis_2xCh_EV} and Figure~\ref{fig:MoCoGAN_vDis_EV} vs \ref{fig:MoCoGAN_vDis_2xCh_EV}). A consistent observation is that the magnitude of the largest eigenvalue tends to reduce throughout training and that this effect is less observable with an increase in discriminator kernel dimensionality.

\begin{figure}[pos=!h]
    \centering
    \begin{subfigure}[t]{0.5\textwidth}
        \centering
        \includegraphics[width=1\linewidth]{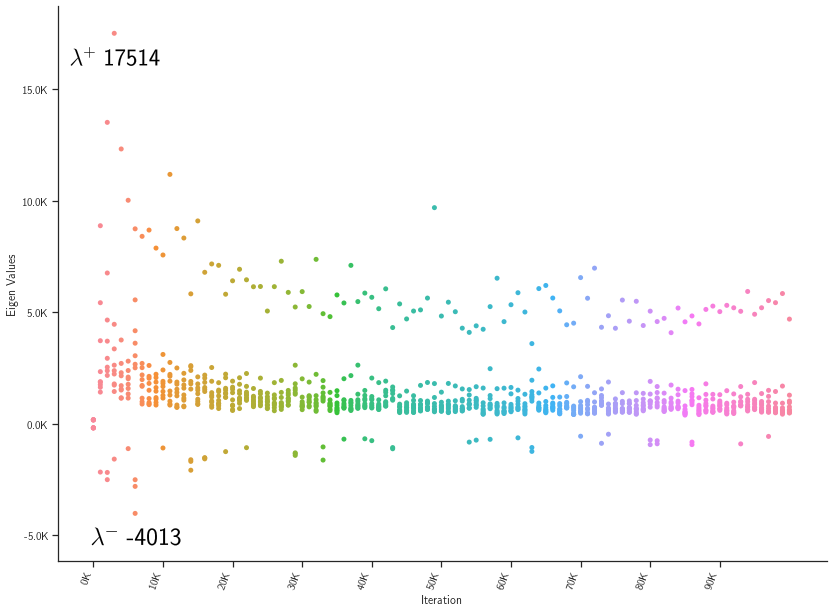}
        \caption{MoCoGAN ksize8 - Image Discriminator}
        \label{fig:MoCoGAN_iDis_EV_ksize8}
    \end{subfigure}%
    \begin{subfigure}[t]{0.5\textwidth}
        \centering
        \includegraphics[width=1\linewidth]{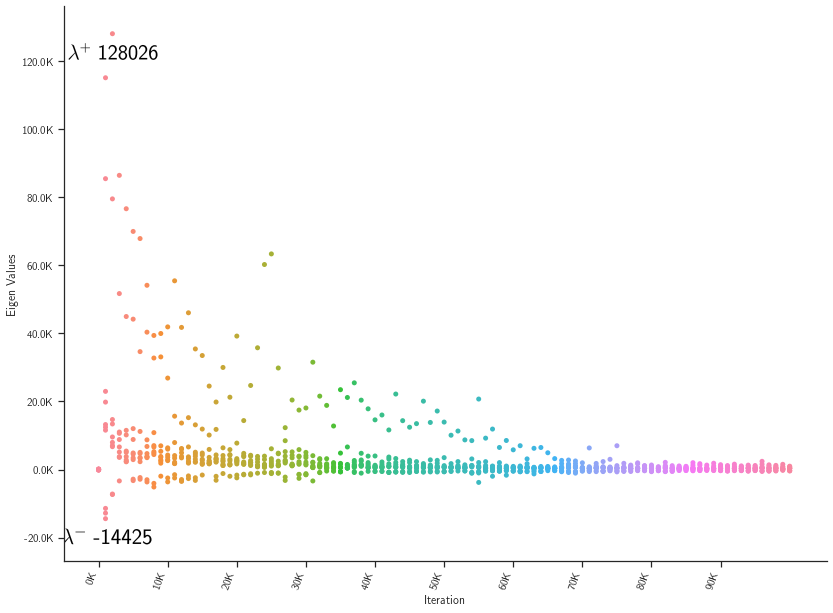}
        \caption{MoCoGAN ksize2 - Video Discriminator}
        \label{fig:MoCoGAN_vDis_EV_ksize2}
    \end{subfigure}%
    \\
    \begin{subfigure}[t]{0.5\textwidth}
        \centering
        \includegraphics[width=1\linewidth]{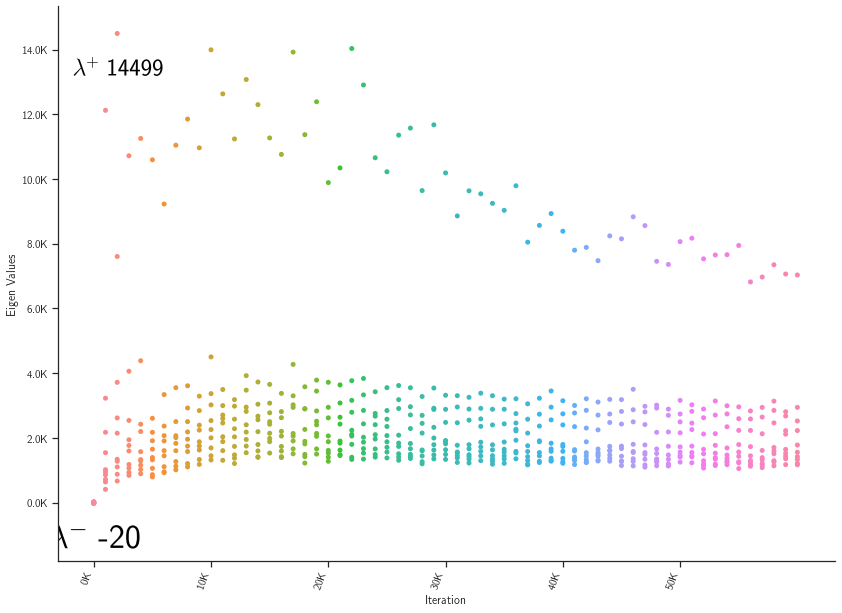}
        \caption{MoCoGAN MUG - Image Discriminator}
        \label{fig:MoCoGAN_iDis_EV_MUG}
    \end{subfigure}%
    \begin{subfigure}[t]{0.5\textwidth}
        \centering
        \includegraphics[width=1\linewidth]{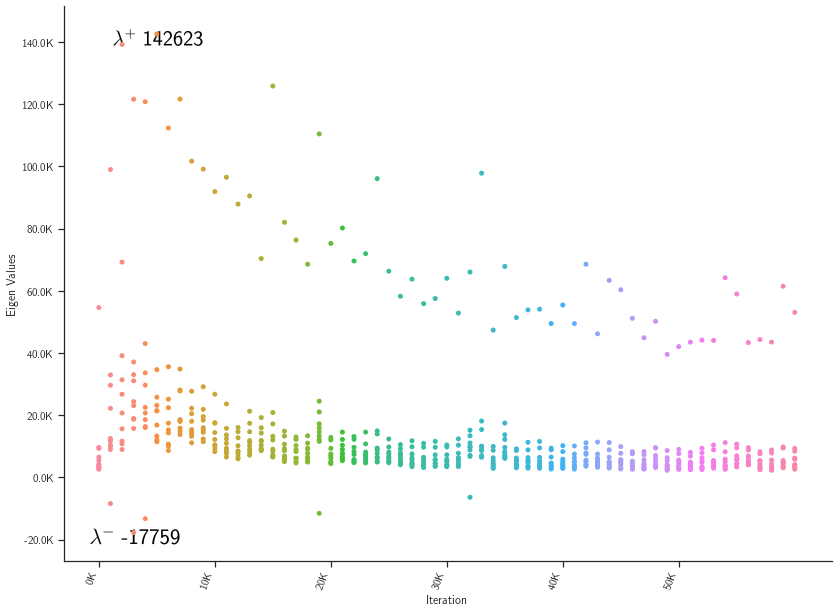}
        \caption{MoCoGAN MUG - Video Discriminator}
        \label{fig:MoCoGAN_vDis_EV_MUG}
    \end{subfigure}%
\caption{The 10 largest eigenvalues of the loss Hessian with respect to the image and video discriminators of the MoCoGAN architecture. (a) shows these values for an image discriminator with kernels that have four times the parameters of the original image discriminator. (b) shows these values for a video discriminator with an eighth the parameters per kernel of the original video discriminator. (c) and (d) show these values for the MoCoGAN discriminators trained on the MUG-FED dataset. $\lambda^{+}$ denotes the largest positive eigenvalue encountered during training and $\lambda^{-}$ the largest negative eigenvalue.} 
    \label{fig:MoCoGAN_Dis_EV2}
\end{figure}

\paragraph{Kernel Complexity vs Kernel Dimensionality:}
It could be said that the observations in Figure~\ref{fig:MoCoGAN_Dis_EV} are possibly an artifact of kernel complexity rather than dimensionality. But increasing the parameter complexity of the image discriminator such that it matches that of the video discriminator leads to little deterioration in the loss landscape of the image discriminator (Figure~\ref{fig:MoCoGAN_iDis_EV_ksize8} vs Figure~\ref{fig:MoCoGAN_vDis_EV}). Instead we observe that even with an increase in kernel complexity, the loss landscapes induced by the image discriminators are more similar to each other in terms of curvature than to those induced by the video discriminator (Figure~\ref{fig:MoCoGAN_iDis_EV}, \ref{fig:MoCoGAN_iDis_2xCh_EV}, and \ref{fig:MoCoGAN_iDis_EV_ksize8} vs Figure~\ref{fig:MoCoGAN_vDis_EV}). Furthermore, Figure~\ref{fig:MoCoGAN_vDis_EV_ksize2} shows that before collapse, a video discriminator with half the kernel complexity of an image discriminator induces a similar loss landscape to that of the original MoCoGAN video discriminator ( Figure~\ref{fig:MoCoGAN_vDis_EV} vs  Figure~\ref{fig:MoCoGAN_vDis_EV_ksize2}) . 

\newpage

\paragraph{Dataset Complexity vs Kernel Dimensionality:} An argument could be made that the dataset complexity affects the curvature of the loss landscape. We analyse the loss surface of the MoCoGAN model trained on two different datasets with very different mode characteristics and scene dynamics, UCF-101 and MUG-FED. UCF-101 includes 101 different classes with different inter and intra class variability for each video when compared to MUG-FED. MUG-FED is a dataset with a fixed background and different faces under controlled lighting making different facial expressions. UCF-101 is a significantly more complex dataset than MUG-FED and our analysis shows that indeed, the loss landscape is affected by the dataset, resulting in landscapes with different characteristics (see Figure~\ref{fig:MoCoGAN_iDis_EV} vs \ref{fig:MoCoGAN_iDis_EV_MUG} and Figure~\ref{fig:MoCoGAN_vDis_EV} vs \ref{fig:MoCoGAN_vDis_EV_MUG}). But we still observe that the curvature of this landscape is primarily dictated by the kernel dimensionality resulting in similar curvature magnitude profiles for models trained on either dataset (see Figure~\ref{fig:MoCoGAN_iDis_EV}, \ref{fig:MoCoGAN_vDis_EV} vs Figure~\ref{fig:MoCoGAN_iDis_EV_MUG}, \ref{fig:MoCoGAN_vDis_EV_MUG}). In particular, we observe the same order of magnitude difference between the eigenvalues of the loss Hessian for 2D image discriminators when compared against their 3D video counterparts.

\subsubsection{Explaining the Dual Video Discriminator}
\label{sec:dvd_explain}
Stochastic gradient decent (SGD) requires a smooth loss landscape for stable optimisation. An ill-conditioned Hessian alludes to directions of high curvature in this landscape that lead to instabilities in the optimization of typical neural networks \cite{saarinen1993ill, Martens_HessFree}. The GAN optimisation process is itself also highly unstable, exhibiting rotational mechanics and cyclical dynamics that are detrimental to convergence \cite{balduzzi18aMechanics}, more so if the true data distribution is concentrated on a lower dimensional manifold as is likely in video \cite{Nagarajan17_StableGD,mescheder18aConverge}. 

Figure~\ref{fig:MoCoGAN_Dis_EV} shows that the optimisation landscape induced by video GAN discriminators has significant pathologies. For the MoCoGAN discriminators, the highest Hessian eigenvalue observed is at times up to an order of magnitude larger than the next leading eigenvalue. This behaviour is observed at the early stages of training for the 2D image discriminator but these extreme outliers exist throughout training for the 3D video discriminator. More importantly, Figure \ref{fig:MoCoGAN_Dis_EV} shows that these pathologies are made worse as the kernel dimensionality of the discriminator increases. The eigenvalues of the loss Hessian induced by the video discriminator are altogether almost an order of magnitude larger than those of the image discriminator.  Figure~\ref{fig:MoCoGAN_Dis_EV2} shows that this is irrespective of dataset and kernel parameter complexity. 
Altogether, these observations help to explain the emergence of dual 2D image and 3D video discriminators in models such as MoCoGAN \cite{Tulyakov0YK18mocogan}. SGD and its derivatives face a bigger challenge optimising the loss landscape induced by a 3D discriminator when compared to that of a 2D discriminator. As a result, the 2D discriminator likely improves performance by providing a better image-level gradient signal for the generator. The observations in Table~\ref{table:mocogan_ablate}, row 6, where increasing the number of parameters for the 3D video discriminator did not lead to a significant boost in performance lend further support to this hypothesis.

\subsection{TGAN Discriminator}
\begin{figure}[pos=!h]
    \centering
    \begin{subfigure}[t]{0.5\textwidth}
        \centering
        \includegraphics[width=1\linewidth]{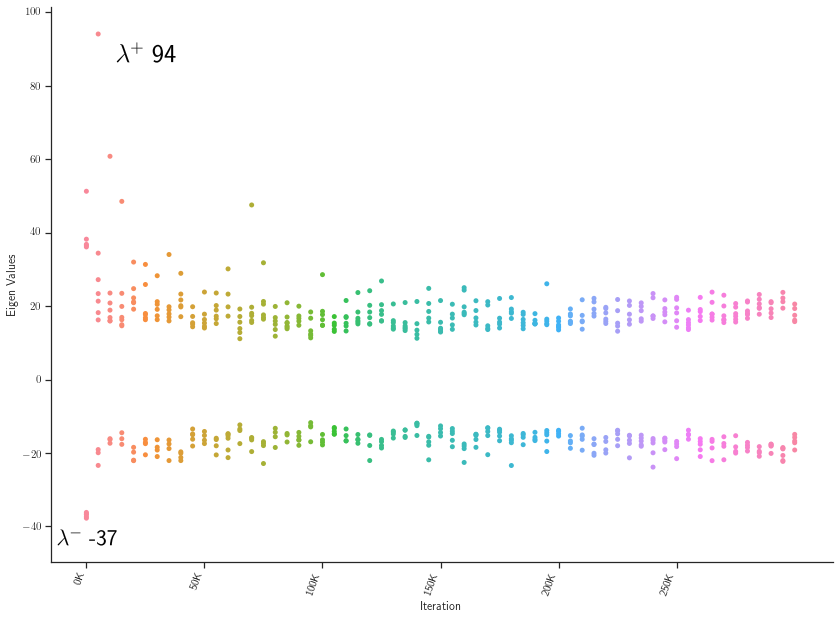}
        \caption{TGAN - SVC}
        \label{fig:TGAN_SVC_EV}
    \end{subfigure}%
    \begin{subfigure}[t]{0.5\textwidth}
        \centering
        \includegraphics[width=1\linewidth]{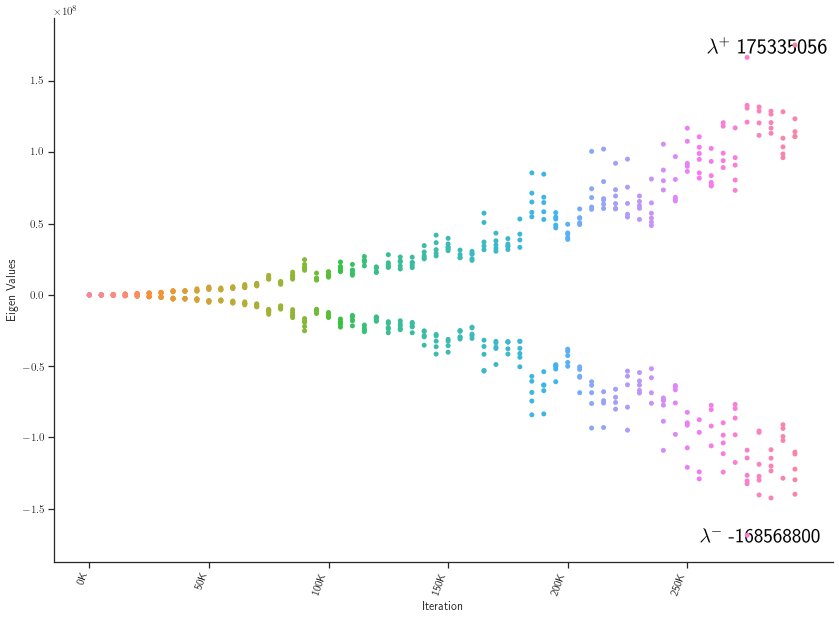}
        \caption{TGAN - Normal}
        \label{fig:TGAN_Normal_EV}
    \end{subfigure}
\caption{The 10 leading eigenvalues of the loss Hessian with respect to the TGAN discriminator, with and without SVC applied.  $\lambda^{+}$ denotes the largest positive eigenvalue encountered during training and $\lambda^{-}$ the largest negative eigenvalue.} 
    \label{fig:TGAN_EV}
\end{figure}

In the previous section, we established that naive application of 3D discriminators induces loss landscapes with high curvature when compared to that induced by 2D discriminators. TGAN \cite{SaitoMS17temporal} utilizes a single 3D video discriminator and manages to match the performance of dual discriminator models like MoCoGAN. In our replication study (see Table~\ref{table:ldvd_is64}); TGAN consistently outperforms MoCoGAN. Inspired by Wasserstein GAN \cite{ArjovskyCB17wasserstein}, TGAN proposes clamping of the spectral norm of the discriminator to a maximum of one in order to stabilize training. This is achieved in practice via Singular Value Clipping (SVC) of the weight matrices such that all singular values are equal to or less than one, enforcing a 1-Lipschitz constraint on the video discriminator. This clipping is applied every $n$ iterations during training and the TGAN authors demonstrate that applying SVC leads to stable training and significantly better performance. We analyse the loss landscape induced by SVC and observe how it impacts model performance. The results are presented in Table~\ref{table:tgan_model} and Figure~\ref{fig:TGAN_EV}.

\begin{table}[width=\linewidth,cols=5,pos=!h]
\caption{TGAN Model}
\label{table:tgan_model}
\begin{subtable}{.59\linewidth}
\caption{TGAN Reproduction}
\label{table:tgan_ablate}
    \begin{tabular*}{0.98\linewidth}{@{} LCC@{} }
    \toprule
    Model  & IS $\uparrow$ & FID $\downarrow$\\
    \midrule
    TGAN - SVC                                                      & 11.93 $\pm$ .08 & 9127.80 $\pm$ 13.77 \\
    TGAN - SVC - Original \cite{SaitoMS17temporal}                  & 11.85 $\pm$ .07 &  \\
    \midrule
    TGAN - Normal                                                   & 8.98  $\pm$ .06 & 10093.02 $\pm$ 11.06 \\
    TGAN - Normal - Original \cite{SaitoMS17temporal}               & 9.18  $\pm$ .11 &  \\
    \midrule
    TGAN - SVC - 2xTime                                & 13.28 $\pm$ .09 & 8797.95 $\pm$ 10.39    \\
    \bottomrule
    \end{tabular*}
\end{subtable}%
\begin{subtable}{.41\linewidth}
        \caption{TGAN Discriminator}
        \label{table:tgan_disc_arch}
        \begin{tabular*}{0.98\linewidth}{@{} LL@{} }
        \toprule
        Layer & Block Configuration \\
        \midrule
        Input & 16 $\times$ height $\times$ width $\times$ 3  \\
        \midrule
        c0 & Conv3D-(N64, K4, S2, P1), LReLU\\
        c1 & Conv3D-(N128, K4, S2, P1), BN, LReLU\\
        c2 & Conv3D-(N256, K4, S2, P1), BN, LReLU\\
        c3 & Conv3D-(N512, K4, S2, P1), BN, LReLU\\
        c4 & Conv2D-(N1, K4, S1, P0,)\\
        \bottomrule
    \end{tabular*}
\end{subtable}
\end{table}

Figure~\ref{fig:TGAN_EV} provides an interesting insight into the TGAN discriminator, especially when contrasted against the MoCoGAN video discriminator (see Figure~\ref{fig:MoCoGAN_vDis_EV}). The TGAN discriminator induces a loss landscape filled with saddle points, characterised by the symmetry between the positive and negative eigenvalues of the loss Hessian. Without SVC, we observe that the curvature of the loss landscape becomes more extreme throughout training (Figure~\ref{fig:TGAN_Normal_EV}). With SVC applied, we observe a comparatively smooth loss landscape with a better conditioned Hessian. This leads to more stable training dynamics and better performance as shown in Table~\ref{table:tgan_ablate}. 

We also observe from results presented in row 2 of Table~\ref{table:mocogan_ablate} and the last row of Table~\ref{table:tgan_ablate}, that temporal subsampling of video frames has a significant impact on model performance. When temporal subsampling of frames is experimentally controlled for, it appears that TGAN significantly outperforms MoCoGAN according to both FID and IS.

\section{Lower-Dimensional Video Discriminators}
\label{sec:ldvd}
We have established that good discriminator performance is promoted by a smooth loss landscape, which is well understood for neural networks. We have shown how enforcing a 1-Lipschitz constraint on the discriminator can smooth the loss landscape. We have also demonstrated that there is a strong correlation between the conditioning number of the Hessian and the kernel dimensionality of the discriminator. This opens up an interesting direction in terms of discriminator architecture design. 

\begin{itemize}
    \item \textit{"Do video discriminators require 3D kernels?"}
\end{itemize}

In addition to the higher curvature optimisation landscapes induced by higher dimensional discriminators, there are other disadvantages to using higher dimensional kernels such as an increase in computation and memory costs. In this section, we propose solutions to these issues by exploiting the insights gained from Section~\ref{sec:disc_arch}.

Most generators for video GANs utilize kernels with a maximum dimensionality of two \cite{SaitoMS17temporal,Tulyakov0YK18mocogan, saito2018tganv2}, but all discriminators currently incorporate 3D kernels.  Thus, we explore the possibility that it may be possible to capture temporal dynamics using a more compressed kernel representation since locally, most information useful for video discrimination may lie on a lower dimensional manifold. This hypothesis is supported by results from related domains such as video recognition, where \cite{TranWTRLP18r2plus1d,FeichtenhoferPW16spatiotemporal,SunJYS15factorized,QiuYM17pseudo3D,XieSHTM18rethinking,linGHtsm} have successfully removed 3D kernels from classification models without compromising model performance. In most cases, performance has improved and our observations from Section~\ref{sec:disc_arch} provide a possible explanation for this phenomenon, a better conditioned loss Hessian. Similarly, we seek to replace 3D video GAN discriminators with lower dimensional approximations, resulting in memory and computational efficiency gains as well as better performance due to more stable training dynamics.

We now introduce a family of Lower Dimensional Video Discriminators for Generative Adversarial Networks (LDVD-GANs). These discriminators are characterised by having a maximal kernel dimension that is lower than the ambient dimension of the data modality they are applied to.

\subsection{Factorized Convolutions}
\label{sec:factorized_conv}
Decomposing 3D convolution kernels into 2D and/or 1D is an area of active research interest in video recognition and understanding \cite{TranWTRLP18r2plus1d,FeichtenhoferPW16spatiotemporal,SunJYS15factorized,QiuYM17pseudo3D,XieSHTM18rethinking}. This process can formally be defined as: 
\begin{equation}\label{eq:factorizedK}
    \textbf{K}^{h,w,t} = \textbf{A}^{h,w} \otimes \textbf{b}^{t}. \qquad \qquad \text{where} \hspace{1em} \textbf{K} \in \mathbb{R}^{k_h \times k_w \times k_t}, \textbf{A} \in \mathbb{R}^{k_h \times k_w} , \textbf{b} \in \mathbb{R}^{k_t},
\end{equation}
are convolution kernels and $\otimes$ denotes the Kronecker product. $\textbf{K}$ is from the subset of kernels that can be factorized into $\textbf{A}$ and $\textbf{b}$ as shown in Eq.~\ref{eq:factorizedK}. A convolution over a feature map $\textbf{F} \in \mathbb{R}^{f_h \times f_w \times f_t}$ can then be defined as:
\begin{equation}\label{eq:factorizedConv}
    \textbf{F}^{i+1} = (\textbf{F}^i_{f_h,f_w} \incircbin{*} \textbf{A})_{f_t} \incircbin{*} \textbf{b}
\end{equation}

where $\incircbin{*}$ denotes the convolution operation\footnote{This includes all operations associated with deep learning convolutions; e.g. padding, dilation, etc} and the subscripts denote the dimensions over which it is applied. 

The R(2+1)D model \cite{TranWTRLP18r2plus1d} applies the factorization sequentially, as shown in Eq.~\ref{eq:factorizedConv}, while the S3D-G \cite{XieSHTM18rethinking} model combines it with a feature gating mechanism. The Pseudo-3D (P3D) family of models \cite{QiuYM17pseudo3D} explores different orderings of spatial and temporal convolutions within the bottleneck block of a 2D residual network. The $F{_{ST}}CN$ model \cite{SunJYS15factorized} splits the network in half, applying spatial convolutions to the first half and temporal convolutions to the down-stream features.  In the two-stream literature, ST-ResNet \cite{FeichtenhoferPW16spatiotemporal} applies a temporal-spatial-temporal factorization within its bottleneck blocks combined with inter-stream residual paths. All of these models present impressive results when compared to their 3D counterparts within their application domains.

The factorized convolution applied in our discriminators is similar to the one shown in Eq.~\ref{eq:factorizedConv}, with the addition of an activation layer between the spatial and temporal convolutions. That is, our factorised convolution is of the form:
\begin{equation}\label{eq:factorizedConvAct}
    \textbf{F}^{i+1} = \text{LReLU}(\textbf{F}^i_{f_h,f_w} \incircbin{*} \textbf{A})_{f_t} \incircbin{*} \textbf{b} 
\end{equation}

\begin{table}[width=\linewidth,cols=7,pos=!h]
\centering
\caption{Factorized discriminators}
\label{table:disc_factorized}
\begin{subtable}{.5\linewidth}
\caption{MoCoGAN - Factorized}
\label{table:mocogan_factorized}
    \begin{tabular*}{0.98\linewidth}{@{} LLL@{} }
        \hline 
        \textbf{Layer}       & \textbf{Configuration}  & \textbf{Block Operations}    \\ 
        \hline
        \textbf{c0$_{h,w}$}  & K(1,4,4),S(1,2,2),P(0,1,1),ch64      & c0$_{h,w}$, LReLU            \\ 
        \textbf{c0$_{t}$}    & K(4,1,1),S1,P(1,0,0),ch64      & c0$_t$, LReLU                \\ 
        \hline
        \textbf{c1$_{h,w}$}  & K(1,4,4),S(1,2,2),P(0,1,1),ch128     & c1$_{h,w}$, LReLU            \\ 
        \textbf{c1$_{t}$}    & K(4,1,1),S1,P(1,0,0),ch128     & c1$_t$, BN, LReLU          \\
        \hline
        
        \textbf{c2$_{h,w}$}  & K(1,4,4),S(1,2,2),P(0,1,1),ch256     & c2$_{h,w}$, LReLU            \\ 
        \textbf{c2$_{t}$}    & K(4,1,1),S1,P(1,0,0),ch256     & c2$_t$, BN, LReLU          \\
        
        
        \hline
        \textbf{c3$_{h,w}$}  & K(1,4,4),S(1,2,2),P(0,1,1),ch256     & c4$_{h,w}$, LReLU            \\ 
        \textbf{c3$_{t}$}    & K(4,1,1),S1,P(1,0,0),ch1       & c4$_t$  \\    
        \hline 
    \end{tabular*}
\end{subtable}%
\begin{subtable}{.5\linewidth}
\caption{TGAN - Factorized}
\label{table:tgan_factorized}
    \begin{tabular*}{0.98\linewidth}{@{} LLL@{} }
        \hline 
        \textbf{Layer}       & \textbf{Configuration}  & \textbf{Block Operations}    \\ 
        \hline
        \textbf{c0$_{h,w}$}  & K(1,4,4),S(1,2,2),P(0,1,1),ch64      & c0$_{h,w}$, LReLU            \\ 
        \textbf{c0$_{t}$}    & K(4,1,1),S(2,1,1),P(1,0,0),ch64      & c0$_t$, LReLU                \\ 
        \hline
        \textbf{c1$_{h,w}$}  & K(1,4,4),S(1,2,2),P(0,1,1),ch128     & c1$_{h,w}$, LReLU            \\ 
        \textbf{c1$_{t}$}    & K(4,1,1),S(2,1,1),P(1,0,0),ch128     & c1$_t$, BN, LReLU          \\
        \hline
        
        \textbf{c2$_{h,w}$}  & K(1,4,4),S(1,2,2),P(0,1,1),ch256     & c2$_{h,w}$, LReLU            \\ 
        \textbf{c2$_{t}$}    & K(4,1,1),S(2,1,1),P(1,0,0),ch256     & c2$_t$, BN, LReLU          \\
        
        \hline
        \textbf{c3$_{h,w}$}  & K(1,4,4),S(1,2,2),P(0,1,1),ch512     & c3$_{h,w}$, LReLU            \\ 
        \textbf{c3$_{t}$}    & K(4,1,1),S(2,1,1),P(1,0,0),ch512     & c3$_t$, BN, LReLU    \\
        
        \hline
        \textbf{c4$_{h,w}$}  & K(4,4),S1,P0,ch1                     & c4$_{h,w}$, LReLU            \\ 
        \hline 
    \end{tabular*}
\end{subtable}
\end{table}

The factorized MoCoGAN and TGAN discriminator architectures are presented in Table~\ref{table:disc_factorized}. They are identical to their original counterparts with the exception that all the 3D convolution kernels are factorized.

\subsection{Temporal Shift Module}
The Temporal Shift Module (TSM) \cite{linGHtsm}, entirely forgoes 3D and/or 1D kernels. Instead, it adapts a purely 2D network for video processing by shifting a portion of channels temporally. This allows for an increase in the temporal receptive field of each layer controlled by the temporal shift distance and the layer depth. 

For MoCoGAN, we apply the TSM to the MoCoGAN image discriminator architecture and use that as the sole discriminitive function during training. For TGAN, we first replace all 3D convolution layers with their 2D counterparts and then interleave convolution operations with shifting operations via the use of TSMs.

Temporal shifting is only applied between intermediate layers and the temporal shift distance in either direction is a single time-step applied to a quarter of the channels for each direction as in \cite{linGHtsm}.

\section{Experiments}
\label{sec:experiments}
It is important to note that we do not do any hyper-parameter tuning, nor do we deviate from the experimental setups of the original TGAN and MoCoGAN experiments. Our sole focus in these experiments is to apply the insights gained from Section~\ref{sec:disc_arch} and demonstrate the efficacy of using Lower Dimensional Video Discriminators (LDVDs) for video generation. In doing so we also improve on the state-of-the-art for video generation and provide a significantly more efficient architecture for high-resolution video generation, competitive with state-of-the-art multi-gpu models. Crucially, we achieve all this by showing that our proposed lower dimensional discriminators can double the performance of previously published models and set state-of-the-art results using only a single GPU. 

\subsection{Factorized Convolutions}

\begin{figure}[pos=!h]
    \centering
    \begin{subfigure}[t]{0.5\textwidth}
        \centering
        \includegraphics[width=\linewidth]{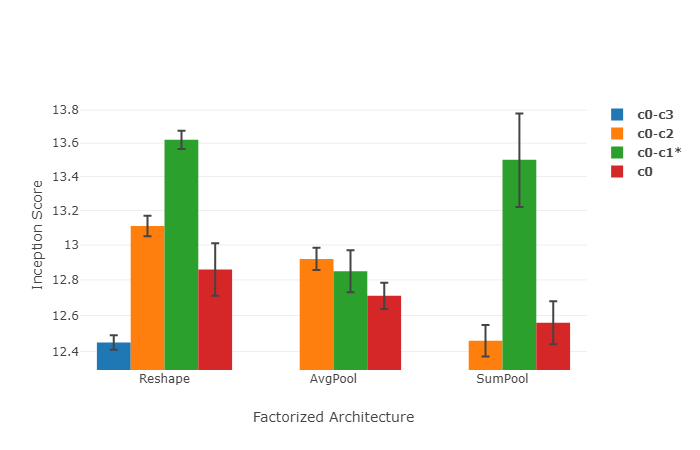}
        \caption{Pooling strategy vs Performance}
        \label{fig:isFconv64arch}
    \end{subfigure}%
    ~
    \begin{subfigure}[t]{0.5\textwidth}
        \centering
        \includegraphics[width=\linewidth]{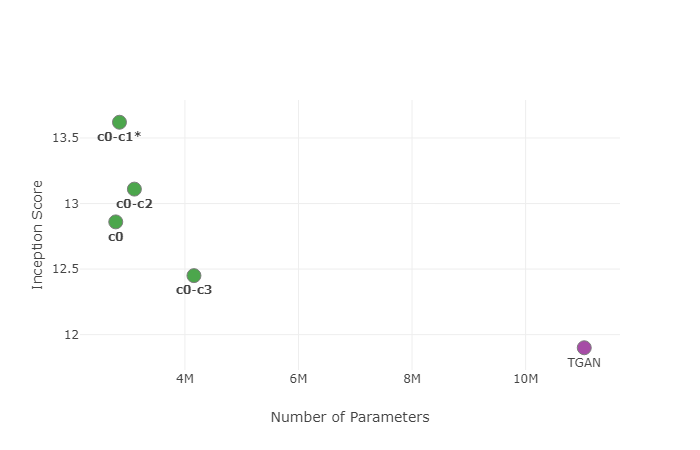}
        \caption{Parameter Efficiency vs Performance}
        \label{fig:isFconv64par}
    \end{subfigure}
    \caption{Performance of factorized TGAN discriminator architectures on 64x64 UCF-101 video generation}%
    \label{fig:tgan_isFconv64}
\end{figure}

We explore the space of lower dimensional video architectures induced by our formulation in Section \ref{sec:ldvd}. We aim to gauge how much dimension factorization affects baseline performance when using a lower dimensional video discriminator. As such, we benchmark lower dimensional discriminators that are factorized to different degrees. 
Our proposed LDVD architectures in Table~\ref{table:disc_factorized} apply factorized convolutions for all 3D convolution layers in their respective video discriminator architectures. For the MoCoGAN discriminator (Table~\ref{table:mocoganDis}), factorizing layer c0 to c3 corresponds to the factorised discriminator architecture shown in  Table~\ref{table:mocogan_factorized} whose performance is presented in the final row of Table~\ref{table:mocogan_factorized_results}. The same applies for the TGAN discriminator (Table~\ref{table:tgan_disc_arch}), where factorizing layer c0 to c3 corresponds to the factorised discriminator architecture shown in Table~\ref{table:tgan_factorized} whose performance is presented in the final row of Table~\ref{table:tgan_factorized_results}. We denote the different discriminators by the layers at which factorization is applied. c0 denotes factorization of the first 3D convolution layer, c0-c1 denotes factorization of the first and second convolution layers, so on and so forth. All other non-factorized convolution layers are restricted to using 2D convolution kernels. The performance of models trained with these discriminators is presented in Table~\ref{table:disc_factorized_results}. The TGAN discriminator has an additional 2D layer appended to it. We explore how temporal aggregation methods applied to its inputs affect model performance. The results are presented in Figure~\ref{fig:tgan_isFconv64}.

\begin{table}[width=\linewidth,cols=8,pos=!h]
\centering
\caption{Performance on the UCF-101 dataset for discriminators with factorization applied to a varying number of convolution layers}
\label{table:disc_factorized_results}
\begin{subtable}{.5\linewidth}
    \caption{TGAN - Factorized Convolutions}
    \label{table:tgan_factorized_results}  
    \begin{tabular*}{0.98\linewidth}{@{} lccr@{}}
    \hline 
    Model                                       & Inception Score$\uparrow$ &  FID$\downarrow$                &       Params        \\
    \hline
    TGAN \cite{SaitoMS17temporal}               & 11.85 $\pm$ .07           &                                &       11M           \\
    TGAN Ours                                   & 11.93 $\pm$ .08           &             9128 $\pm$ 14          &       11M           \\
    \hline
    Layer: c0                                 & \textbf{12.86 $\pm$ .15}  &      \textbf{9031 $\pm$ 2}            &       2.8M          \\
    Layer: c0-c1                              & \textbf{13.62 $\pm$ .06}  &      \textbf{8943 $\pm$ 4}               &       2.8M          \\
    Layer: c0-c2                              & \textbf{13.11 $\pm$ .06}  &      \textbf{8989 $\pm$ 4}   &       3.1M          \\
    Layer: c0-c3                              & \textbf{12.45 $\pm$ .04}  &      \textbf{9082 $\pm$ 1}   &       4.2M          \\
    \hline 
    \end{tabular*}
\end{subtable}%
\begin{subtable}{.5\linewidth}
    \caption{MoCoGAN - Factorized Convolutions}
    \label{table:mocogan_factorized_results}
    \begin{tabular*}{0.98\linewidth}{@{} lccr@{}}
    \hline 
    Model                                       & Inception Score$\uparrow$  &   FID$\downarrow$     &    Params               \\
    \hline
    MoCoGAN \cite{Tulyakov0YK18mocogan}         & 12.42 $\pm$ .03  &                               &       3.3M              \\
    MoCoGAN Ours                                & 11.58 $\pm$ .04  &       9485 $\pm$ 15                &       3.3M              \\
    \hline
    Layer: c0                                 &  9.71 $\pm$ .05  &        9886 $\pm$ 13           &       0.7M              \\
    Layer: c0-c1                              &  \textbf{11.60 $\pm$ .05}  &       \textbf{9424 $\pm$ 12}           &       0.7M              \\
    Layer: c0-c2                              &  10.56 $\pm$ .10  &       9518 $\pm$ 6            &       1M                \\
    Layer: c0-c3            &    10.20 $\pm$ .09  &     9694 $\pm$ 3   &       1M              \\
    \hline 
    \end{tabular*}
\end{subtable}
\end{table}

Table~\ref{table:disc_factorized_results} and Figure~\ref{fig:tgan_isFconv64} provide for several interesting observations; the first being that factorized LDVDs can outperform their 3D counterparts using a fraction of their parameters. In particular for the TGAN discriminator, performance improves in every case. The best performing factorized TGAN discriminator boosts the IS by around 15\% and consequently the state-of-the-art for 64x64 video generation by around 10\%. In comparison, we only observe moderate performance improvements for the MoCoGAN model and suspect that this is likely due to a bottlenecked discriminator. We denote the best performing factorized TGAN and MoCoGAN models as TGAN-F and MoCoGAN-F respectively. These are the c0-c1 discriminators from Table~\ref{table:disc_factorized_results}.

\paragraph{Improving MoCoGAN Performance:} In Section~\ref{sec:disc_arch} we demonstrate three ways of encouraging smooth loss landscapes; increasing the number of discriminator parameters, using an LDVD and directly enforcing that all layers in the discriminator are 1-Lipschitz continuous.  The TGAN model applies Singular Value Clipping (SVC) to stabilize training and its factorized discriminators benefit from it. We explore enforcing a 1-Lipschitz constraint on the MoCoGAN-F discriminator via spectral normalization \cite{MiyatoKKY18spectral} and observe that it improves performance to an IS of 12.33 $\pm$ .09 and an FID of 9069 $\pm$ 4. 

\paragraph{Temporal Resolution:} An interesting observation about the MoCoGAN-F and TGAN-F discriminators is that they only process temporal information in their initial couple of layers. Factorised discriminators with more capacity to process temporal information further downstream and over a longer temporal receptive field do not perform as well. 

\subsection{Temporal Shift Module}
Table~\ref{table:disc_tsm} summarizes experiments from our exploration of the temporal shifting strategy. As in the original TSM models from \citet{linGHtsm}, single-step temporal shifting in each direction is applied  to a quarter of feature maps. Our discriminator naming convention is similar to that in the previous section, whereby ci-ck denotes a discriminator with temporal shifting applied after layers i through k. All discriminators are entirely 2D in nature with temporal shifting used to capture and merge temporal information between intermediate layers.

\begin{table}[width=\linewidth,cols=8,pos=!h]
\centering
\caption{Performance on the UCF-101 dataset for discriminators using the Temporal Shift Modules at different layer depths}
\label{table:disc_tsm}
\begin{subtable}{.5\linewidth}
    \caption{TGAN - Temporal Shifting}
    \label{table:tgan_tsm}  
    \begin{tabular*}{0.98\linewidth}{@{} lccr@{}}
    \hline 
    Model                                       & Inception Score$\uparrow$  &  TRV      &    Params            \\
    \hline
    TGAN \cite{SaitoMS17temporal}               & 11.85 $\pm$ .07  &                22               &       11M             \\
    TGAN Ours                                   & 11.93 $\pm$ .08  &                22               &       11M            \\
    \hline
    Layer: c0                                 & 11.76 $\pm$ .06  &                3                &       2.8M              \\
    Layer: c0-c1                              & \textbf{12.11 $\pm$ .12}  &                5                &       2.8M              \\
    Layer: c0-c2                              & \textbf{12.32 $\pm$ .07}  &                7                &       2.8M              \\
    \hline 
    \end{tabular*}
\end{subtable}%
\begin{subtable}{.5\linewidth}
    \caption{MoCoGAN - Temporal Shifting}
    \label{table:mocogan_tsm}
    \begin{tabular*}{0.98\linewidth}{@{} lccr@{}}
    \hline 
    Model                                       & Inception Score$\uparrow$  &          TRV          &    Params            \\
    \hline
    MoCoGAN \cite{Tulyakov0YK18mocogan}         & 12.42 $\pm$ .03  &                22               &       3.3M             \\
    MoCoGAN Ours                                & 11.58 $\pm$ .04  &                22               &       3.3M              \\
    \hline
    Layer: c0                                 &  8.90 $\pm$ .05  &                3                &       0.7M              \\
    Layer: c0-c1                              &  9.82 $\pm$ .07  &                5                &       0.7M              \\
    Layer: c0-c2                              &   9.60 $\pm$ .02  &                7                &       0.7M              \\
    \hline 
    \end{tabular*}
\end{subtable}
\end{table}

The temporal shift module (TSM) allows for video discriminators with the same number of parameters as traditional image discriminators. Additionally, the parameter cost of the discriminator is constant regardless of the size of the temporal receptive field. This can be seen in Table~\ref{table:tgan_tsm}, where a 2D discriminator improves the performance of the TGAN model by around 10\%, through depth-wise regulation of the Temporal Receptive Field (TRV). We observe a drop in performance for the MoCoGAN model (Table~\ref{table:mocogan_tsm}) and attribute it to the discriminator not being 1-Lipschitz continuous. We denote the best performing discriminators for TGAN and MoCoGAN from Table~\ref{table:disc_tsm}, TGAN-TSM and MoCoGAN-TSM respectively.

When comparing temporal shifting to factorized convolutions, we observe that the temporal processing carried out by the 1D kernels in factorized convolutions result in significant improvements in performance for negligible cost in terms of memory and computation. The shifting strategy is comparatively expensive, often increasing training times by 10-40\% depending on the number of shifting operations carried out.

\subsection{Comparison with State-of-the-Art}
\label{sec:sota}

\subsubsection{A note on the IS of UCF-101 Videos}
An accurate comparison against previous work requires that a distinction be made between low and high resolution video generation. This is because metrics such as the inception score (IS) are known to be sensitive to image resolution \cite{brockDS2018large,BORJI201941}. In video, the additional temporal dimension is unconstrained and requires some procedure for sub-sequence selection. Additionally, data augmentation methods such as frame sub-subsampling (i.e. skip every $n$ frames) or cropping are applied to some published models but not others. We benchmark the IS of the UCF-101 \textit{'trainlist01'} dataset used in the video GAN literature under different conditions. Since dataset videos have more frames than the evaluation networks 16 frame temporal resolution, we can derive a rough standard deviation by repeating the evaluation process four times with randomly sampled 16 frame sub-sequences. These results are shown in Table~\ref{table:true_is}. 

\begin{table}[pos=!h]
\centering
\caption{Inception Score of the UCF-101 training dataset under different conditions}
\label{table:true_is}
    \begin{tabular}{lcccc}
    \hline 
    Inception Score         &  Resolution                       &  2x Subsampling &  Random Video Reversal  &       Crop    \\
    \hline
     59.61 $\pm$ .21 &  16 $\times$ 64 $\times$ 64 $\times$ 3   &                 &                         &    center     \\
    63.69 $\pm$ .32  &  16 $\times$ 64 $\times$ 64 $\times$ 3   &  \checkmark     &                         &    center     \\
   	63.58 $\pm$ .20  &  16 $\times$ 64 $\times$ 64 $\times$ 3   &  \checkmark     &   \checkmark            &    center     \\
    61.83 $\pm$ .21  &  16 $\times$ 64 $\times$ 64 $\times$ 3   &  \checkmark     &   \checkmark            &    random     \\
    \hline
    90.78 $\pm$ .21  &  16 $\times$ 128 $\times$ 128 $\times$ 3 &                 &                         &    center     \\
    91.20 $\pm$ .14  &  16 $\times$ 128 $\times$ 128 $\times$ 3 &  \checkmark     &                         &    center     \\
    90.55 $\pm$ .10  &  16 $\times$ 128 $\times$ 128 $\times$ 3 &  \checkmark     &   \checkmark            &    center     \\
    89.73 $\pm$ .20  &  16 $\times$ 128 $\times$ 128 $\times$ 3 &  \checkmark     &   \checkmark            &    random     \\
    \hline 
    \end{tabular}
\end{table}

\paragraph{Dataset Normalization:}We observed that the IS is highly sensitive to the mean normalisation file used during evaluation and as a result maintain the use of the mean normalisation file provided by the TGAN authors \cite{SaitoMS17temporal}. We note that our 128 resolution results in Table~\ref{table:true_is} are close to the true data IS of 83.18 published in \cite{AcharyaHPG2018towards}. 

\subsubsection{Lower Resolution Generation}

\begin{table}[pos=!h]
\centering
\caption{Performance on the 64x64 UCF-101 video generation benchmark}
\label{table:ldvd_is64}
    \begin{tabular}{lccc}
    \hline 
    Model                                           & Inception Score $\uparrow$             & FID $\downarrow$ &     Parameter Reduction $\uparrow$    \\
    \hline
    VGAN \cite{VondrickPT16generating} (2016)       & 8.31 $\pm$ .09                        &                 &                           \\
    TGAN \cite{SaitoMS17temporal} (2017)           & 11.85 $\pm$ .07                       &                 &                           \\
    MoCoGAN \cite{Tulyakov0YK18mocogan} (2018)     & 12.42 $\pm$ .03                       &                 &                           \\
    
    \hline
    TGAN (our reproduction)                            & 11.93 $\pm$ .08                      & 9127.80 $\pm$ 13.77           &                        \\
    TGAN-TSM                                        & 12.32 $\pm$ .07                   &   9796.68 $\pm$ 2.29          &           74.93\%     \\
    TGAN-F                                       & \textbf{13.62 $\pm$ .06}             &   \textbf{8942.63 $\pm$ 3.72} &   74.19\%          \\
    \hline
    MoCoGAN  (our reproduction)                          & 11.58 $\pm$ .04                      & 9485.34 $\pm$ 14.61           &                        \\
    MoCoGAN-TSM                                     & 9.82 $\pm$ .07                    &  10608.66 $\pm$ 15.67         &   79.99\%            \\

    MoCoGAN-F                                    & 11.60 $\pm$ .05                       &   9424.01 $\pm$ 12.16        &           69.61\%         \\
    
    MoCoGAN-F + SN                               & 12.33 $\pm$ .09                       &   9069.11 $\pm$ 3.97         &           69.61\%         \\
    MoCoGAN + TGAN-F Discriminator               & \textbf{12.53 $\pm$ .01}              & \textbf{9038.42 $\pm$ 9.21}  &           15.16\%         \\

    \hline 
    \end{tabular}
\end{table}

Table~\ref{table:ldvd_is64} presents results for the best performing TSM and factorized discriminators against the state-of-the-art models for 64x64 video generation.

Both TGAN-TSM and TGAN-F outperform the original TGAN architecture using a quarter of the the parameters of the original discriminator. Furthermore, TGAN-F sets a new state-of-the-art result for 64x64 video generation and outperforms complex higher resolution models such as ProVGAN (see Table.~\ref{table:tgan_is128}) without exploiting temporal subsampling, a technique which we have shown to significantly boost performance (see the second row of Table~\ref{table:mocogan_ablate} and the last row of Table~\ref{table:tgan_ablate}). Next, we explore higher resolution video generation with this architecture.

\subsubsection{Higher Resolution Generation}
The computation and memory gains achieved by reducing the maximum kernel dimension of the discriminator allows for the TGAN model to be scaled up to higher resolutions without issue, even on a single GPU system. Our high-resolution video generation model is based on our best performing low-resolution model, TGAN-F, with appropriate modifications made to support higher resolutions. 

\begin{table}[pos=!h]
\centering
\caption{Performance on the UCF-101 benchmark for high-resolution video generation}
\label{table:tgan_is128}
    \begin{tabular}{lcccccr}
    \hline 
    Model                                                   &   Resolution                      &  Batch Size   &     Compute   &   IS$\uparrow$    &     FID$\downarrow$                   & Params  \\
    \hline
    ProVGAN \cite{AcharyaHPG2018towards} (2018)             &   32$\times$256$\times$256        &               &    Multi-GPU  &     13.59                 &                           &    \\
    ProVGAN + SWGAN \cite{AcharyaHPG2018towards} (2018)     &   32$\times$256$\times$256        &               &   Multi-GPU   &    14.56                  &                           &   \\
    TGANv2  \cite{saito2018tganv2} (2018)                   &   16$\times$192$\times$192        &     64        &    4 GPUs     & 22.70 $\pm$ .19           &                           &  200M \\
    \hline
    TGAN-F                                                  &   16$\times$128$\times$128        &     32        &   1 GPU       & 16.85 $\pm$ .04  & 8797 $\pm$ 7   &  16M \\
    TGAN-F + 4xTemporalCh                                   &   16$\times$128$\times$128        &     32        &   1 GPU       & 17.72 $\pm$ .20  & 8361 $\pm$ 11   &   27M \\
    TGAN-F + 2xImageCh                                      &   16$\times$128$\times$128        &     32        &   1 GPU       & 20.35 $\pm$ .23  & \textbf{7817 $\pm$ 10}   &   25M \\
    TGAN-F + 2xTime                                         &   16$\times$128$\times$128        &     32        &   1 GPU       & 17.23 $\pm$ .15  & 8444 $\pm$ 18   &   16M \\
    \hline
    TGAN-F + All                                          &   16$\times$128$\times$128        &     32        &   1 GPU       & \textbf{22.91 $\pm$ .19}        &  8016 $\pm$ 17       &   70M \\
    \hline 
    \end{tabular}
\end{table}

Table~\ref{table:tgan_is128} presents results for TGAN-F benchmarked on the task of 128x128 video generation. TGAN-F (+ 4xTemporalCh) corresponds to quadrupling the number of channels in the temporal frame generator, TGAN-F (+ 2xImageCh) corresponds to doubling the number of channels in the TGAN-F image generator, TGAN-F (+ 2xTime) corresponds to doubling the temporal receptive field by subsampling a longer video and TGAN-F (+ All) corresponds to applying all the above modifications to the TGAN-F architecture. 

The first observation is that TGAN-F trained on a single GPU outperforms all single-GPU models by at least 15\%.  The second observation is that TGAN-F is significantly more parameter efficient than previous video GAN architectures. This consequently enables it to be more memory efficient, enabling training with batch sizes of up to 32 and at resolutions as high as 16x128x128 on a single GPU. Another observation is that TGAN-F (+ All) almost doubles the performance of the original TGAN model while using the same generator architecture, hyper-parameters and hardware constraints. The last observation is that TGAN-F (+ All) on a single gpu provides for state-of-the-art results while using a fraction of the parameters and compute of multi-GPU VGAN models. This demonstrates the efficacy and efficiency of LDVD-GANs like TGAN-F, and shows its superior performance when compared to the original TGAN model and many other video GAN models.

\begin{figure}[pos=!h]
    \centering
    \begin{subfigure}[t]{0.5\textwidth}
        \centering
        \includegraphics[width=1\linewidth]{figs/result_tgan_vdis_eigen_spectra_top10.png}
        \caption{TGAN - SVC}
        \label{fig:TGAN_EV_comp}
    \end{subfigure}%
    \begin{subfigure}[t]{0.5\textwidth}
        \centering
        \includegraphics[width=1\linewidth]{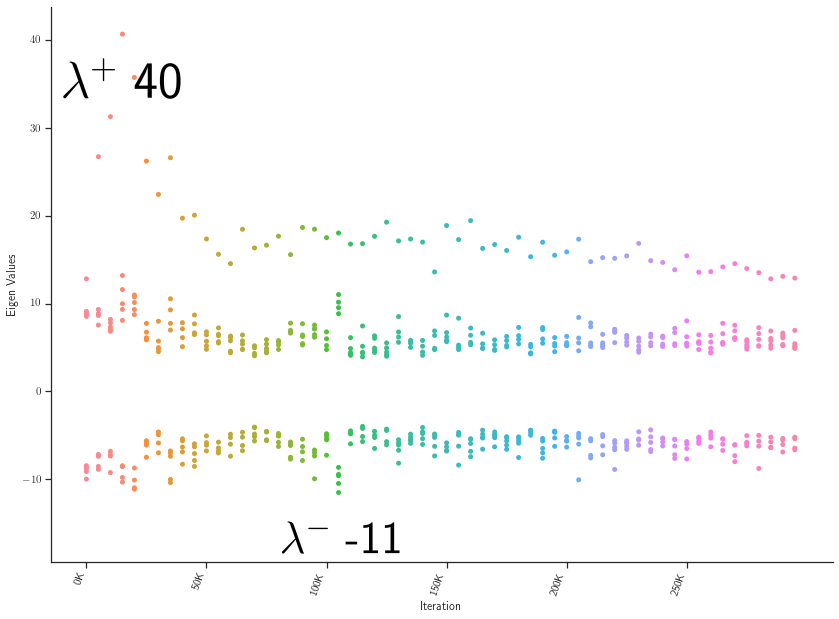}
        \caption{TGAN - F}
        \label{fig:TGAN_F_EV_comp}
    \end{subfigure}
\caption{The 10 leading eigenvalues of the Hessian with respect to the TGAN and TGAN-F discriminator.  $\lambda^{+}$ denotes the largest positive eigenvalue encountered during training and $\lambda^{-}$ the largest negative eigenvalue.} 
    \label{fig:TGAN_EV_Comp}
\end{figure}

\paragraph{TGAN-F Loss Landscape:} Hessian analysis of the loss landscape induced by the TGAN-F discriminator during optimisation shows that the curvature of this space is more than halved when compared to that of the original TGAN discriminator (see Figure~\ref{fig:TGAN_F_EV_comp} vs \ref{fig:TGAN_EV_comp}). 
The smoother loss landscape induced by the lower dimensional discriminator helps to explain the improved performance of the TGAN-F model when compared to its original higher dimensional counterpart, TGAN. 

\subsubsection{Qualitative Results}

\begin{figure}[pos=!h]
    \centering
    \includegraphics[width=1\linewidth]{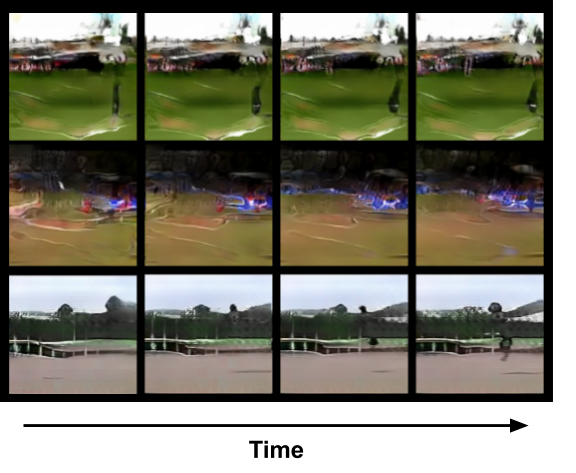}
    
    \caption{Frames from videos generated by TGAN-F trained on UCF-101 at
128$\times$128 resolutions. Top: Zoom into scene, Middle: Pan left to right, Bottom: Rotate around center of focus.} 
    \label{fig:UCF101_ldvd_128}
\end{figure}

Figure~\ref{fig:MUG_ldvd_64_256} shows frames from TGAN-F models trained on MUG-FED at different resolutions. Figure~\ref{fig:UCF101_ldvd_128} shows selected high-resolution samples from the same model trained on the UCF-101 dataset. We observe learned camera zooming and panning motions. Full resolution random samples and other qualitative results are available in the supplementary material. \footnote{Supplementary Material: \url{https://drive.google.com/drive/folders/1J9gjS2HRTwoADQVqVoMbs5pBtO7rOGF6}}

\begin{figure}[pos=!h]
    \centering
    \begin{subfigure}[t]{0.5\textwidth}
        \centering
        \includegraphics[width=\linewidth]{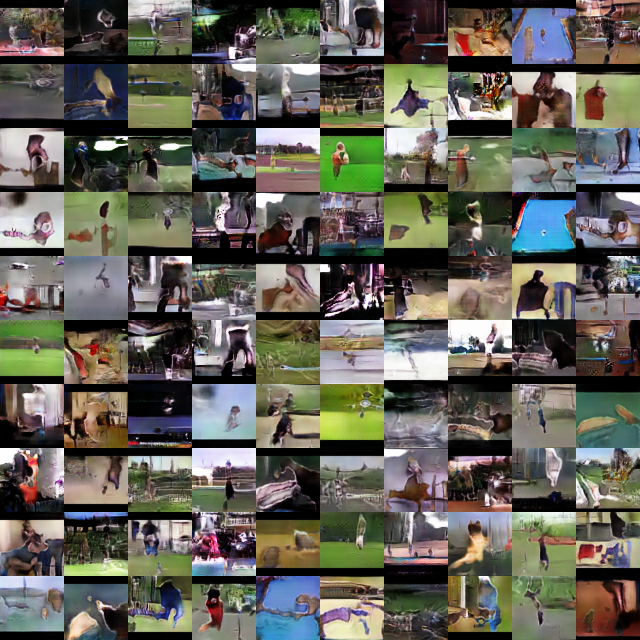}
        \caption{MoCoGAN}
        \label{fig:mocogan_samples_ucf101_64}
    \end{subfigure}%
    ~
    \begin{subfigure}[t]{0.5\textwidth}
        \centering
        \includegraphics[width=\linewidth]{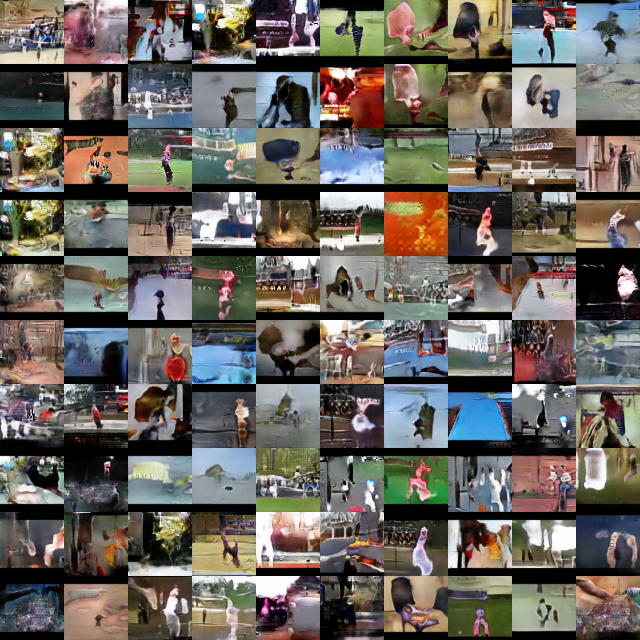}
        \caption{MoCoGAN - F with SN}
        \label{fig:mocoganf_samples_ucf101_64}
    \end{subfigure}%
    \\
    \begin{subfigure}[t]{0.5\textwidth}
        \centering
        \includegraphics[width=\linewidth]{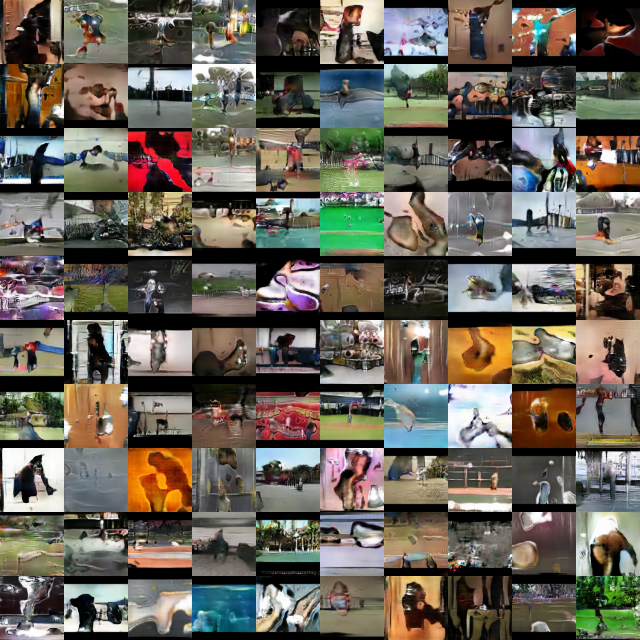}
        \caption{TGAN}
        \label{fig:tgan_samples_ucf101_64}
    \end{subfigure}%
    ~
    \begin{subfigure}[t]{0.5\textwidth}
        \centering
        \includegraphics[width=\linewidth]{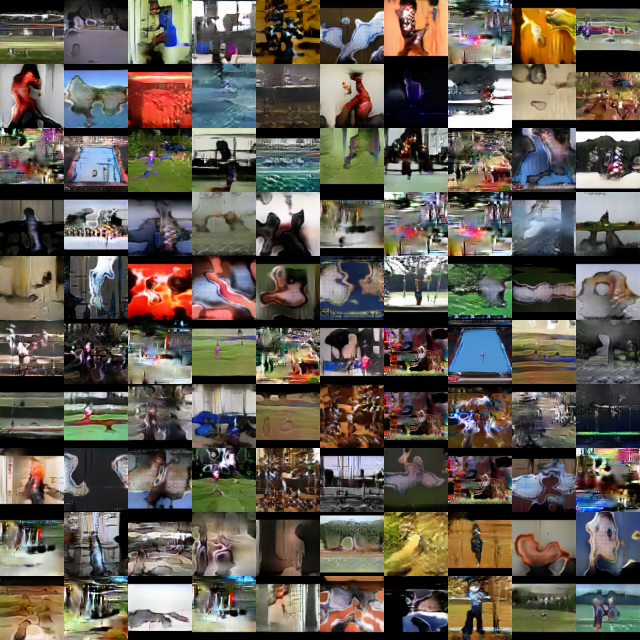}
        \caption{TGAN - F}
        \label{fig:tganf_samples_ucf101_64}
    \end{subfigure}
    \caption{The first frame of video samples for each of the models on 64x64 UCF-101. Full samples available in the supplementary material}%
    \label{fig:ucf101_samples_64}
\end{figure}

\begin{table}[pos=!h]
\caption{Human evaluation of the quality and diversity of samples generated by different models trained on the MUG-FED dataset.}
\label{table:human_eval}
    \begin{tabular}{lccc}
    \hline 
    Model Comparison      &     Quality (\%)   &  Diversity (\%)  \\
    \hline
    MoCoGAN/TGAN          &      33.0/67.0     &    50.0/50.0   \\
    TGAN/TGAN-F           &      41.6/58.4     &    50.0/50.0       \\
    MoCoGAN/TGAN-F        &      25.0/75.0     &    16.6/83.4   \\
    \hline 
    \end{tabular}
\end{table}

\section{Conclusion}
\label{sec:end}
The field of image generation has enjoyed significant advances in recent years, and our work aims at taking a step towards doing the same for video generation. Specifically, we study the properties of video discriminator architectures and find that higher dimensional video discriminators induce a loss landscape with relatively higher curvature. As a result, we question the utility of 3D kernels in video GAN models and empirically demonstrate that they are not required for the video generation problem as it is currently framed. Our design proceeds by replacing 3D kernels with lower dimensional approximations, and our proposed lower dimensional discriminators, improve the performance of video GAN generators they are applied to. As a result, we demonstrate performance that is competitive with the state-of-the-art for both single and multi-gpu video generation; in both low-resolution and high-resolution video generation settings. 

We carried out a wide range of experiments across two generator models and many more discriminator architectures. We summarise the successful results of this investigation in Section~\ref{sec:disc_arch} and Section~\ref{sec:experiments}. These experiments demonstrate that the curvature of the loss landscape for video GAN discriminators increases with kernel dimensionality. We also uncover guiding principles to limit this behaviour; mainly avoiding 3D kernels, but also enforcing a 1-Lipschitz discriminator and increasing the number of parameters in a model (Section~\ref{sec:ldvd}). Based on these principles, we propose a family of lower dimensional video discriminator architectures that provide for efficient but powerful video GAN models. Subsequently, we explore one such lower dimensional discriminator architecture, TGAN-F, resulting in state-of-the-art performance for a single-gpu model (Section~\ref{sec:sota}).\\

\clearpage
\section{Bibliography}

\bibliographystyle{cas-model2-names}

\bibliography{cas-refs}

\end{document}